\documentclass[runningheads,a4paper]{llncs}
\usepackage{bm}
\usepackage{amssymb}
\usepackage{graphicx}
\usepackage{url}
\urldef{\mailsa}\path|{karamjit.singh, puneet.a, gautam.shroff}@tcs.com|    
\usepackage{nicefrac}
\usepackage{amsmath}
\usepackage{verbatim}
\usepackage{url}
\urldef{\mailsa}\path|{karamjit.singh, puneet.a, gautam.shroff}@tcs.com|

\begin{document}

\mainmatter  

\title{Warranty Cost Estimation Using Bayesian Networks}

\titlerunning{Warranty Cost Estimation}

%
%
\author{Karamjit Singh
\and Puneet Agarwal\and Gautam Shroff}
\authorrunning{Warranty Cost Estimation }

%
\institute{TCS Research\\
\mailsa\\}

%

\toctitle{Lecture Notes in Computer Science}
\tocauthor{Authors' Instructions}
\maketitle

\begin{abstract}
\emph{All multi-component product manufacturing companies face the problem of warranty cost estimation. Failure rate analysis of components plays a key role in this problem. Data source used for failure rate analysis has traditionally been past failure data of components. However, failure rate analysis can be improved by means of fusion of additional information, such as symptoms observed during after-sale service of the product, geographical information (hilly or plains areas), and information from tele-diagnostic analytics. In this paper, we propose an approach, which learns dependency between part-failures and symptoms gleaned from such diverse sources of information, to predict expected number of failures with better accuracy. We also indicate how the optimum warranty period can be computed. We demonstrate, through empirical results, that our method can improve the warranty cost estimates significantly.}
\end{abstract}

\section{Introduction and Motivation}
All multi-component product manufacturing companies attempt to estimate the cost of warranty support, in order to draw their annual budget. 
However, it is hard to estimate this cost accurately, and there is a penalty associated with both under-estimation and over-estimation.
The under-estimation can cause shortage of failed components in the market, leading to customer dissatisfaction and can diminish brand-name of the company. While, over-estimation has high opportunity cost from the financial perspective. 
It is therefore very important for manufacturing companies to predict warranty cost accurately. The warranty cost estimates usually depends on three factors: 1) number of warrant-able items, 2) failure rate analysis, and 3) cost per failure, where failure rate analysis plays a significant role in estimation of the warranty cost \cite{5}.

Most of the existing studies use past \textit{part-failure data} of components to learn the failure rate parameters corresponding to
probability distributions like Weibull, log-normal, etc.\cite{2}. However, more information is available in enterprises, which can be fused
with the past part-failure data in order to improve the accuracy of estimation of these parameters, which results in better estimates of failures. The warranty cost is the product of part replacement cost and estimated failures in the warranty period. So better estimate of failures leads to a better estimate of warranty cost. For example in an automotive company, service stations data and tele-diagnostics data is available (we describe
these datasets in Section \ref{sec:overview}). These datasets, when fused with past part-failure data, can work as a leading indicator of
part failure and thereby help improve the warranty cost estimates.

Fusion of such data from multiple sources is a non-trivial problem because occurrence of symptoms in vehicle service data or in tele-diagnostics data does not entail part-failure. However, there is a conditional dependence of such symptoms on the part-failure rate. This degree of conditional dependence may change over time with sales of newer models of the product and change in number of product units sold, as a result the conditional dependence cannot be learned statically once, and used later. 
Therefore, in this paper, we propose a model for fusion of data from multiple sources using
an approach based on \textit{Bayesian Network}, which can learn the conditional dependency of symptom data on part-failure rate, resulting in improvement of failure estimates, hence better warranty cost. In addition we indicate how the optimum warranty period can be computed. We also substantiate our claims through empirical results on simulated data, which has been inspired from real-life data.

In this paper, we begin by giving an overview of our approach, after a brief description of various datasets involved, in the immediate next section. After that in section 2, we present formal definitions of the data and other terminologies used in rest of the paper, and in section 3 we give an introduction to Bayesian networks and how parameter estimation can be done in Bayesian Networks. We start describing our approach by introducing the proposed Bayesian Network, which is used to learn dependencies between part failure and symptom data, in Section 4. Later, in section 5, we present our approach of warranty cost estimation in detail. We summarize the empirical results in Section 6 and discussed the related work in Section 7. Finally, we conclude with a brief discussion on future work in Section 8.

\subsection{Overview}\label{sec:overview}
To give an overview of our approach to predict warranty cost estimation in multi-component products with better accuracy, we use the example of vehicles as multi-component products in this section.

Consumers take their vehicles to a service station to get it examined by service engineers, who record DTCs (diagnosed trouble codes) \cite{3} for every trouble symptom observed in the vehicle.
The vehicles are taken to the service station either because of failure of a component or on regular service routine.
Further, in many cases before failure of a part, a trouble symptom occurs in the vehicle, and such trouble symptoms are also observed during service of a vehicle.
Although, these trouble symptoms are fixed in the vehicle by the service engineers through incidental preventive maintenance, this information can also be used as a leading indicator for part-failure.

Further, usually these DTCs can be observed only when the vehicle is examined by service engineers at the service station. However, the
modern vehicles \cite{6} are equipped with sensors, which not only observe the state of various components but also transmit this
information, intermittently or regularly, to a central hub for offline analysis. Researchers are trying to build algorithms to predict
part-failure from such sensor data \cite{6}, \cite{24} and these methods are referred as \textit{tele-diagnostics}. In this paper we
consider that if the tele-diagnostics data were available, how can we fused it with the past part-failure data, in order to improve the
warranty cost estimates. In contrast, the traditional approach is to find expected number of failure using past part-failure data \cite{2}
only, we present a novel approach which captures the dependencies in a Bayesian network and uses these dependencies to predict expected
number of future part failures with better accuracy.

Traditionally, failure rate parameter(s) estimation has been done component-wise i.e. considering whole component as one unit. However, as each component can further be divided into sub-components or more granular levels, i.e., part wise. Considering the fact that each part can have different failure rate, learning the failure rate parameter(s) on granular level rather than on component level can enhance the accuracy of prediction. 

In our approach we model a simple Bayesian network $(F \rightarrow I \rightarrow S)$, following a process that when a part fails $(F)$, a trouble symptom in terms of DTC $(I)$ occurs in the vehicle, and the drivers takes the vehicle to service station where the DTC is diagnosed or observed $(S)$. 
Here, a DTC may occur before part-failure, but it is not observed until the vehicle is taken to the service station. A DTC will definitely occur when a part in the vehicle fails. 
A vehicle could be taken to the service station because of vehicle service routine or due to failure of a part. So the occurrence time of DTC can be much early than the observed time of DTC. we model the $F$ node using Weibull distribution, and the $I$ and $S$ nodes using Gaussian distribution. The parameters of these distributions defines the dependency between the nodes $F$, $I$, and $S$. The Bayesian network to learn dependencies between $F$, $I$, and $S$ for each part and and related DTC is presented in section \ref{modelBN} and the detail approach to learn the dependencies and further use these dependencies  to predict expected number of failures is presented in section \ref{ExfailureBN}.

\section{Data Description}\label{dataDesc}
We consider data for the time duration $t_{1}$ to $t_{2}$ for $n$ products indexed from 1 to $n$, where each product has $m$ parts. The data is obtained from three sources: Part failure data, Service Records, and tele-diagnostics data. The datasets from the three sources are:
\begin{description}
\item $\bullet$ \textbf{Part Failure Data:} In part failure data, we have a variable $p_{j,i}$ : \textit{number of cycles} at which part $P_{j}$ fails in $i^{th}$ product, where $i \in \{1, 2,..., n\}$ and $j \in \{1, 2,..., m\}$. These cycles can be time-to-failure (hours, days, seconds) for a product, miles for which a vehicle is driven, etc.
 
 \item $\bullet$ \textbf{DTC Occurrence Data}: We assume each part $P_{j}$ is associated with a set of DTCs s.t.  $P_{j}$ $\leftarrow$ \{$D_{j,1}$, $D_{j,2}$,.., $D_{j,r}$\}, where $D_{j,k}$ is a DTC associated with the part $P_{j}$,  $k \in \{1, 2,..., r\}$ and no two DTCs can be associted with same part. When a part fails, all the DTCs associated with it occur. One or more DTCs associated with the part may occur before the part fails. In this data, we have variable $d_{j,k}$: number of cycles at which DTC $D_{j,k}$ associated with part P$_{j}$ occurs first time in the tele-diagnostic data.
 
 \item $\bullet$ \textbf{DTC Observed Data}: In this data, we have a variable $s_{j,k}$: number of cycles at which DTC $D_{j,k}$ associated with part P$_{j}$ is observed first time in service records.
\end{description}
Next, we define some terms which will be used in the rest of the paper. 
\subsection{Definitions}
\begin{description}
\item $\bullet$ \textbf{C$_{\textbf{j}}$}: This set contains index of products in which part $P_{j}$ fails for the first time in the time-interval [$t_{1}$, $t_{2}$] where j = 1 to m.  
\begin{center}
C$_{j}$ = \{ $i$: index of a product in which part $P_{j}$ fails in the time interval [$t_{1}$, $t_{2}$] \}
\end{center}
Let the cardinality of the set $C_{j}$ is $n_{j}$, i.e., $\mid C_{j}\mid$ = $n_{j}$, where 1$\leq$ $n_{j}$ $\leq$ n.
 \item $\bullet$ \textbf{Fail$_{\textbf{j}}$}: This set contains number of cycles at which part $P_{j}$ fails for each product in $C_{j}$.
 \begin{equation}
   Fail_{j} = \{ p_{j,i}: i \in C_{j} \}, \mid Fail_{j}\mid = n_{j}\}
 \end{equation}
\item $\bullet$ \textbf{Ind$_{\textbf{j,k}}$}: This set contains number of cycles at which DTC $D_{j,k}$ associated with part $P_{j}$, occurs first time for each product in $C_{j}$. So the set $Ind_{j,k}$ will contain $d_{j,k,i}$  $\forall$ $p_{j,i}$ $\in$ $Fail_{j}$.
\begin{equation}
 Ind_{j,k} = \{ d_{j,k,i}: i \in C_{j} \}, \mid Ind_{j,k}\mid = n_{j}\}
\end{equation}
 \item $\bullet$ \textbf{Serv$_{\textbf{j,k}}$}: This set contains number of cycles at which DTC $D_{k}$ associated with part $P_{j}$ observed first time for each product in $C_{j}$. $Serv_{j,k}$ will contain $s_{j,k,i}$  $\forall$ $p_{j,i}$ $\in$ $Fail_{j}$.
 \begin{equation}
  Serv_{j,k} = \{ s_{j,k,i}: i \in C_{j} \}, \mid Serv_{j,k}\mid = n_{j}\}
 \end{equation}
\end{description}
 It is to be noted that $p_{j,i}$, $d_{j,k,i}$, and $s_{j,k,i}$ from the sets $Fail_{j}$, $Ind_{j,k}$, and $Serv_{j,k}$ respectively will always satisfy following relation:
\begin{equation}
  d_{j,k,i} \leq s_{j,k,i} \leq p_{j,i}
\end{equation}

Since we have datasets: part failure, tele-diagnostic, and service records for the time interval $[t_{1}, t_{2}]$, i.e., part which fails after time $t_{2}$ and before $t_{1}$ will not be present in given data. As DTCs associated with part can occur before the part failure. So there can be some parts which fails after $t_{2}$, but there corresponding DTCs could have occurred or/and observed in the time interval $[t_{1}, t_{2}]$.

\begin{description}
\item $\bullet$ \bm{$C^{'}_{\textbf{j,k}}$}: This set contains index of products in which part $P_{j}$ fails first time after time $t_{2}$ but it's associated DTC $D_{j,k}$ occurs as well as observed first time in the time interval $[t_{1}, t_{2}]$. Let $\mid C^{'}_{\textbf{j,k}} \mid = n_{j}^{'}$, where $0 \leq n_{j}^{'} \leq n$.  

\item $\bullet$ \bm{$Ind^{'}_{\textbf{j,k}}$}: This set contains number of cycles at which DTC $D_{j,k}$ associated with part $P_{j}$ occurs first time in the time interval $[t_{1}, t_{2}]$ for all cars in $C_{j,k}^{'}$.
\begin{equation}
Ind_{j,k}^{'} = \{ d_{j,k,i}^{'}: i \in C_{j,k}^{'} \}, \mid Ind_{j,k}^{'} \mid = n_{j}^{'}
\end{equation}
\item $\bullet$ \bm{$Serv^{'}_{\textbf{j,k}}$}: This set contains number of miles at which DTCS $D_{j,k}$ associated with part $P_{j}$ observed first time in the time interval $[t_{1}, t_{2}]$ for all cars in $C_{j,k}^{'}$.
\begin{equation}
Serv_{j,k}^{'} = \{ s_{j,k,i}^{'}: i \in C_{j,k}^{'} \}, \mid Serv_{j,k}^{'} \mid = n_{j}^{'}
\end{equation}
\end{description}
\section{Background}
In this section, we give an introduction to Bayesian Networks and how parameters can be learned using Bayesian Networks. In next section, we will present our Bayesian Network which is used to learn the dependency between part failure and DTCs.

A Bayesian network is a representation of joint probability distribution. This representation consists of two components. The first component G is a directed acyclic graph whose vertices correspond to random variables $X_{1}$, $X_{2}$, ..., $X_{n}$. The second component, $\theta$ describes a conditional distribution for each variable, given its parent G. Together these two components specify a unique distribution on $X_{1}$, $X_{2}$, ..., $X_{n}$. The graph G encodes the independence assumption, by which variable X$_{i}$ is independent of its nondescendents given its parents in G \cite{1}, \cite{9}.

By applying the chain rule of probabilities and properties of conditional independence, and joint probability distribution that satisfies Marcov assumptions can be decomposed into the product form\\
\begin{equation}
P(X_{1}, X_{2},..., X_{n})= \prod_{i=1}^{n} P(X_{i} \mid Pa^{G}(X_i))
\end{equation}
where $Pa^{G}(X_i)$ is the set of parents of $X_i$ in G.

The parameters of Bayesian network can be learned using various methods like Maximum-likelihood estimation, EM, MCMC etc. Different methods to learn the parameters of Bayesian Network can be classified based on two factors: 1) Observability of network, i.e., All nodes are known or not, 2) Bayesian structure is known or not \cite{1}. 
\section{Modeling Dependence Using Bayesian Networks}\label{modelBN}
As discussed in section 1, DTCs which occur before the actual part failure can be used as the leading indicators of part failure. We assume that every part $P_{j}$ and it's associated DTC $D_{j,k}$ has some dependency. Few examples of these  dependencies are 1) DTC $D_{j,k}$ always occur, roughly two months before part failure $P_{k}$, 2) DTC $D_{j,k}$ which occurs before the actual part $P_{k}$ failure, follow some probability distribution.\\
Our goal is to learn the dependency between $D_{j,k}$ and $P_{j}$ $\forall$ j = 1 to m and $\forall$ k = 1 to r.

The Bayesian networks combines the Bayesian probability theory and the notion of conditional independence to represent dependencies among variables \cite{25}. In this section, we present a Bayesian network which can learn the dependencies or patterns between part failure and their associated DTCs.

Given the part failure data, tele-diagnostics data, and service records, the Bayesian network in fig 1 is used to learn the dependency between DTC and part failure. Since each part is associated with set of DTCs ( as explained in section \ref{dataDesc}), this Bayesian network will learn the parameters of dependency between number of cycles at which part $P_{j}$ fails($F_{j}$), number of cycles  at which associated DTC $D_{j,k}$ occurs($I_{j}$), and number of cycles at which associated DTC $D_{j,k}$ is observed($S_{j}$).
The various random variables of the Bayesian network given in fig 1 are explained below. Random variables are denoted by capital letters and the values taken by random variables are denoted by corresponding small letters.
\begin{description}
\item $\bullet$ \textbf{F$_{\textbf{j}}$}: It is defined as number of cycles at which part $P_{j}$ fails first time. It take values from the set $Fail_{j}$ . It follows the Weibull distribution with scale parameter as $\beta_{j}$ and shape parameter as $\alpha_{j}$. 
\begin{equation}
 F_{j} \sim Weibull(\alpha_{j}, \beta_{j})
\end{equation}
\item $\bullet$ $\bm{\alpha_{j}}$: It follows the Uniform distribution with lower limit as 0 and upper limit as $a>0$. 
\begin{equation}
 \alpha_{j} \sim U(0, a).
\end{equation}
\item $\bullet$ $\bm{\beta_{j}}$: It follows the Uniform distribution with lower limit as 0 and upper limit as b. where $b>0$ and $b\in \mathbb{R}$.
\begin{equation}
 \beta_{j} \sim U(0, b). 
\end{equation}
\item $\bullet$ \textbf{I$_{\textbf{j,k}}$}: It is defined as number of cycles at which DTC $D_{j,k}$ associated with part $P_{j}$ 
\textit{occurs} first time. It take values from the set $Ind_{j,k}$. It follows the Normal distribution with mean as $f_{j}-f_{j}\times r_{j,k}$ and standard deviation as $\sigma_{j,k}^{1}$. i.e. 
\begin{equation}
 I_{j,k} \sim \mathcal{N}(f_{j}  - f_{j}\times r_{j,k}, \sigma_{j,k}^{1})
\end{equation}
\item $\bullet$ \textbf{R$_{\textbf{j,k}}$}: It follows Uniform distribution. where $r_{1},r_{2} >0$ and $r_{1}, r_{2} \in \mathbb{R}$
\begin{equation}
 R_{j,k} \sim U(r_{1}, r_{2})
\end{equation}
\item $\bullet$ $\bm{\sigma_{j,k}^{1}}$: It follows the Uniform distribution. where $c_{1} > 0$ and $c_{1} \in \mathbb{R}$
\begin{equation}
  \sigma_{j,k}^{1} \sim U(0, c_{1})
\end{equation}
\item $\bullet$ \textbf{S$_{\textbf{j,k}}$}: It is defined as number of cycles at which DTC $D_{j,k}$ associated with part $P_{j}$, \textit{observed} first time. It takes values from the set $Serv_{j,k}$. It follows the Normal distribution with mean as $(f_{j} - i_{j,k})\times m_{j,k} + i_{j,k}$ and standard deviation as $\sigma_{j,k}^{2}$.
\begin{equation}
  S_{j,k} \sim \mathcal{N}((f_{j} - i_{j,k})\times m_{j,k} + i_{j,k}, \sigma_{j,k}^{2}).
\end{equation}
\item $\bullet$ $\bm{M_{j,k}}$: It follows Uniform distribution.
\begin{equation}
   M_{j,k} \sim U(0, 1)
\end{equation}
\item $\bullet$ $\bm{\sigma_{j,k}^{2}}$: It follows Uniform distribution. where $c_{3}>0$ and $c_{3} \in \mathbb{R}$
\begin{equation}
  \sigma_{j,k}^{2} \sim U(0, c_{3})
\end{equation}
\end{description}

\begin{figure}
\centering
\includegraphics[height=6.2cm]{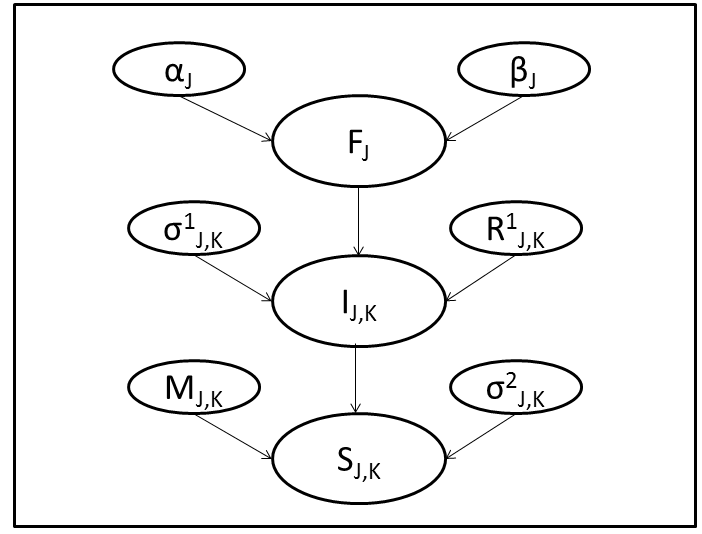}
\caption{Bayesian Network to learn dependency parameters between Part $P_{j}$ and it's associated DTC $D_{j,k}$}
\label{fig 1}
\end{figure}

\subsection{Learning Parameters of Bayesian Network}
Dependencies between part $P_{j}$ and it's associated DTC $D_{j,k}$ i.e. dependency between $F_{j}$, $I_{j,k}$, and $S_{j,k}$ can be captured by learning parameters $R_{j,k}$, $\sigma_{j,k}^{1}$, $M_{j,k}$, $\sigma_{j,k}^{2}$ in our Bayesian network (fig 1). The scale and shape parameters $\beta_{j}$ and $\alpha_{j}$ for the part $P_{j}$ is used to find expected number of failures(explained in \ref{failureParam}. We learn these parameters using the method of MCMC \cite{8} implemented by  Metropolis – Hastings algorithm \cite{12}.
\section{Failure Rate Analysis Using Bayesian Networks}
In this section, we first present an approach to find expected number of failures using the scale and shape parameters of Weibull distribution. Later in this section, we present the detail approach to find failure rate parameters using the Bayesian network presented in section \ref{modelBN}.
\subsection{Expected Number of Failures}\label{failureParam}
Assuming each part $P_{j}$ failure follows Weibull distribution with shape parameter as $\alpha_{j}$ and scale parameter as $\beta_{j}$. Given the scale and shape parameters, we can calculate expected number failures of part $P_{j}$ $\forall$ j = 1 to m. Expected number of failures for n cars (each car has m parts) in time interval $t_{3}$ to $t_{4}$ is given by the equation
\begin{equation}
= \sum_{i=1}^{n} \sum_{j=1}^{m} F_{j}(T)
\end{equation}
where $F_{j}(T)$ is the probabilty that the part will fail in the time interval $t_{3}$ to $t_{4}$.

\subsection{Expected Number of Failures Using Bayesian Networks}\label{failureBN}
Warranty cost estimation is directly affected by the prediction of expected number of failures i.e. failure rate analysis. Estimation of expected number of failures can be improved by incorporating additional information like tele-diagnostic data and service records. The Bayesian network explained in section 4, models the dependency between part failure data, occurrence time of DTC, and observed time of DTC. In this section, we present three cases and approach to find an expected number of failures using Bayesian networks in each case. Figure 2 shows which datasets are used in different cases. In first case, we use only part failure data to do the failure rate analysis. In second case, we use part failure data and service records and in third case, we use part failure data, service records, and sensor data to do failure rate analysis. Comparison of these three cases will presented in section 7. Three cases and their corresponding approaches, used to find expected number of failure of part $P_{j}$ associated with DTC $D_{j,k}$ in each case is presented below:
\begin{figure}
\centering
\includegraphics[height=2.0cm]{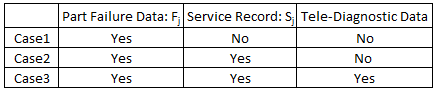}
\caption{Datasets used in different cases}
\label{fig 2}
\end{figure}

\begin{description}
\item \textbf{Case 1}: In this case, we use only part failure data for failure rate analysis. Using the Bayesian network (in section \ref{modelBN}), we learn the parameters $\alpha_{j}$, $\beta_{j}$ for the part $P_{j}$. While learning these parameters, variable $F_{j}$ is considered as known with values taken from the set $Fail_{j}$ and other variables $I_{j,k}$ and $S_{j,k}$, $R_{j,k}$, $\sigma_{j,k}^{1}$, $M_{j,k}$, $\sigma_{j,k}^{2}$ are considered as unknown. Once we learned the $\alpha_{j}$, $\beta_{j}$, we use these parameters to predict expected number of failures for part $P_{j}$ $\forall$ j = 1 to m as explained in section 5.\\

\item \textbf{Case 2}: In this case, we present an approach that incorporates part failure data and service records to find expected number of failures using our Bayesian network (explained in section \ref{modelBN}). Fig 3 explains, which variables are considered as known or unknown during which step i.e. O: observed or N.O: Not observed ( steps are explained below) and in case of known variable, the value of that variable is taken from which set. For example O: $Fail_{j}$ means that variable is considered as known in this step and it's value is taken from the set $Fail_{j}$ and O: step 1 mens that value is known and taken as value learned in step 1 . This approach has four steps which are explained below: 
\begin{description}
\item \textbf{Step 1 - Learning dependency parameters}: In this step, we learn the parameters $R_{j,k}$, $\sigma_{j,k}^{1}$, $M_{j,k}$, $\sigma_{j,k}^{2}$ using our Bayesian network, that captures the dependency between part $P_{j}$ and DTC $D_{j,k}$. While learning these parameters by our Bayesian network( fig 1), variable $F_{j}$ and $S_{j}$ are considered as known with values taken from the sets $Fail_{j}$ and $Serv_{j,k}$ respectively and variable $I_{j}$, $R_{j,k}$, $\sigma_{j,k}^{1}$, $M_{j,k}$, $\sigma_{j,k}^{2}$, $\alpha_{j}$, $\beta_{j}$ are considered as unknown.

\item \textbf{Step 2 - Predicting future failures}: In this step, we predict those failures of part $P_{j}$, in which the part fails after time $t_{2}$ but associated DTC $D_{j,k}$ \textit{occurs} and \textit{observed} in the time interval $[t_{1}, t_{2}]$. To predict these failures, we use the parameters that are learned in previous step. While predicting these failures from our Bayesian network, $R_{j,k}$, $\sigma_{j,k}^{1}$, $M_{j,k}$, and $\sigma_{j,k}^{2}$ considered as known with values learned in previous step, value of $S_{j,k}$ is also considered as known with values taken from the set $Serv^{'}_{j,k}$ (6). All the failures of part $P_{j}$ that are predicted using this step forms a set called $Fail_{j,k}^{'}$.\\

\item \textbf{Step 3 - Learning Weibull parameters}: In this step, we learn the scale $\beta_{j}$ and shape $\alpha$ parameters using our Bayesian network. To learn these parameters, first we take union of two sets $Fail_{j}$ (1) and $Fail_{j,k}^{'}$ (using previous step).
\begin{equation}
Fail_{j,k}^{''} = Fail_{j} \cup Fail_{j,k}^{'}
\end{equation}
\begin{equation}
Serv_{j,k}^{''} = Serv_{j,k} \cup Serv^{'}_{j,k}
\end{equation}
It is to be noted that $Fail_{j} \cap Fail_{j,k}^{'} = \emptyset$ and $Serv_{j,k} \cap Serv^{'}_{j,k} = \emptyset$
While learning scale and shape parameters from our Bayesian network, variables $S_{j,k}$ and $F_{j}$ is considered as known with values taken from the sets $Serv_{j,k}^{''}$ and $Fail_{j,k}^{''}$ respectively and other variables  $I_{j}$, $R_{j,k}$, $\sigma_{j,k}^{1}$, $M_{j,k}$, $\sigma_{j,k}^{2}$, $\alpha_{j}$, $\beta_{j}$ are considered as unknown. Let $\alpha^{new}_{j}$ and $\beta^{new}_{j}$ be the variables learnt in this step.\\

\item \textbf{Step 4 - Finding expected number of failures}: In this step, we use the parameters $\alpha^{new}_{j}$ and $\beta^{new}_{j}$ learned in previous step to predict expected number of failures in the time interval $[t_{3}, t_{4}]$ as explained in section 5.\\
\end{description}

\begin{figure}
\includegraphics[height=1.8cm]{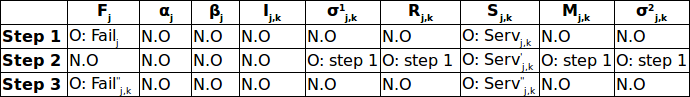}
\caption{Variable in case 2}
\label{fig 3}
\end{figure}

\item \textbf{Case 3}: In this case, approach used to find expected of failures is similar to the approach presented in case 2 with the exception that the variable $I_{j,k}$ is also considered as known along with $F_{j}$ and $Serv_{j,k}$. Fig 4 explains that which variables are considered as known or unknown i.e. O: observed or N.O: Not observed during which step ( steps explained in case 2) and in case of known variable, the value of that variable is taken from what set is given in fig 4. For example in step 1: Learning dependency patterns, variable $F_{j}$, $S_{j,k}$, and $I_{j,k}$ are considered as known i.e O: observed and their values taken from the sets $Fail_{j}$, $Serv_{j}$, and $Ind_{j}$ respectively. 
\end{description}

\begin{figure}
\includegraphics[height=1.8cm]{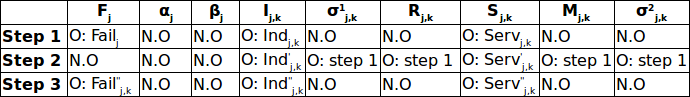}
\caption{Variable in Case 3}
\label{fig 4}
\end{figure}\label{ExfailureBN}
\subsection{Warranty Cost Estimation}\label{CostEstimation}
Warranty cost is the product of part replacement cost and estimated failures in the warranty period. However, there is another cost that incurred to the company in terms customer dissatisfaction called \textit{penality cost}. This penalty cost is for the parts which fails after warranty period and it decreases with time. So the warranty cost $C_{j}$ for the $j^{th}$ part with the warranty period as $w_{j}$ is given below.
\begin{equation}
C_{j} = R_{j}F(w_{j})+ R_{j}be^{-cw_{j}}(1-F(w_{j}))\label{Cost}
\end{equation}
where $b, c>0 \in R$, $F(w_{j})$ corresponds to fraction of parts which fail before $w_{j}$ and $R_{j}$ is the replacement cost for the part $P_{j}$.

The optimal warranty period that minimize the overall cost $C_{j}$ can be easily computed by the method of gradient descent\cite{7} for any specified values of b, c in eq \ref{Cost}.
\section{Experiment and Results}
In this section, we present a comparison of the expected number of failures for the three cases described in Section \ref{failureBN} on simulated data for vehicles. We simulated data corresponding to part failure, sensor data, and service records. We first describe the three simulated datasets, and then present the comparison of three cases (discussed in Section \ref{failureBN} on these datasets. We also present the results of optimal warranty period (discussed in Section \ref{CostEstimation}) 
We used Python module PYMC \cite{11} for learning the dependency parameters of the Bayesian network. We use the learnt parameters to predict the number of part failures in future.

\subsection{Simulated datasets}
We simulated data for 1000 cars, where each car has nine parts. We have cars manufactured over three years 2010, 2011, and 2012. The data for the period 1-Jan-2010 to 31-Dec-2012 is considered for learning the dependency parameters of the Bayesian network. The parameters learnt are then used to predict the number of part failures in the year 2013. The data simulation details for part failure data, sensor data and service records are as follows:
\begin{description}
\item \textbf{Part failure data}: To generate part failure data, we assume that number of miles before failure for a part $P_{j}$ follows Weibull distribution with shape and scale parameters $\alpha_{j}$ and $\beta_{j}$, respectively, where $j \in \{1, 2,..., 9\}$. Number of miles at which $P_j$ fails is generated using inverse transform sampling method \cite{13}.
\item \textbf{Sensor data}: We make the following assumptions to generate sensor data:
\begin{enumerate}
\item Each part is associated with four DTCs and no two parts have any DTC in common.
\item When a part fails, all four DTCs associated with it occur. Also, one or more DTCs associated with the part may occur before the actual part fails.
\item Number of miles at which DTCs associated with a part occur before the actual part failure is generated assuming the probability distribution
 $\mathcal{N}(f - f \times r, \sigma_{1})$ (refer Section \ref{modelBN}), where $f$ is number of miles at which part fails and $r \in (0.1, 0.5)$.
\end{enumerate}
\item \textbf{Service Records}: We assume that a car will go for service: (1) if a part fails in a car, or (2) as soon as the time since last service exceeds six months. Each service record of a car contains DTCs which have occurred since the last service of the car.
\end{description}

\subsection{Observations and Results}
In this section, we compare the results of three cases discussed in Section \ref{failureBN}. We also compare the results with the \textbf{Best Scenario} where we use the same approach to find expected number of failures as explained in Case 3 with the exception that we skip the step 1 of Case 3. Also, in step 2, rather than predicting the failures we form the set $Fail_{j}^{'}$ with the actual future failures as present in the simulated data. Steps 3 and 4 are the same as in the case 3. Best scenario assumes that for the given service records and sensor data, our Bayesian network model can predict actual number of miles for every failure.
 
In Fig 5, first column lists the names of nine parts. In column `Actual parameters', we have shown the scale and shape parameters used to simulate part failure data for each part. We compare these actual parameters with the parameters learned from our model using three different approaches in Case 1, Case 2, and Case 3 (in section 6). Fig 5 shows that the difference in the parameters learned using Case 1 (using only part failure data) and actual parameters is very high for three parts named as E2P, T1P, and T2P. This is due to insufficient number of failures of these parts before 31-Dec-12. i.e.  number of samples in set $Fail_{j}$ for these three parts are not sufficient to learn parameters accurately. However, column `Parameters learned using Case 2' in Fig 5 shows that when we use the approach discussed in Case 2 the values of the parameters for these three parts have improved. Similarly, when we use the approach discussed in Case 3, values of parameters for these three parts further improved. Also, the values of the parameters learned using Case 1 for the parts E1P, E3P, T3P, B1P, B2P, and B3P are close to the actual values. There is small improvement in the parameters learned using Case 2 and Case 3 for the parts E1P, T3P, and B3P. Values of the parameters learned using Case 2 and Case 3 for the parts E3P, B1P, and B2P do not change. This is due to the reason that $Serv_{j,k}^{'} = \emptyset $ and $Ind_{j,k}^{'} = \emptyset$ for these parts (Section \ref{ExfailureBN}). Column `Best Scenario' shows the values of scale and shape parameters learned for four parts using Best scenario case as explained above.

Fig 6 shows the comparison of expected number of failures for the year 2013, calculated for four parts E1P, E2P, T2P, and T1P for the Cases 1-3 and the best scenario. It shows that the difference in actual failures and expected number of failures calculated using Case 1 is high for the parts E2P, T1P, and T2P. This difference reduces when using Case 2, and reduces further when using Case 3.

Fig 7 shows the optimal warranty period for each part calculated using parameters learned in Case 3(as explained in section \ref{CostEstimation}). Values of coefficients b and c (in eq \ref{Cost}) are taken as e (2.71828) and 1/100000 respectively. The value of b is taken in such a way that the penalty for the parts which fails at early stage is higher than the replacement cost of the part and value of c is taken by the assumption that a car runs roughly 100000 miles in three years. The penalty cost is high for small value of w (warranty period) and gets decreases with time. We observe that warranty period is directly proportional to the scale parameter of part. Since we have not made any assumptions about the relative costs of parts, i.e. the $R_i$ (in eq \ref{Cost}), we have not computed the single optimal period as that would depend heavily on this relative cost. However, with the assumption that slower failing parts (higher $\alpha$) cost more, the overall optimal period will be skewed towards the more expensive parts.

\begin{figure}
\centering
\includegraphics[height=4.2cm]{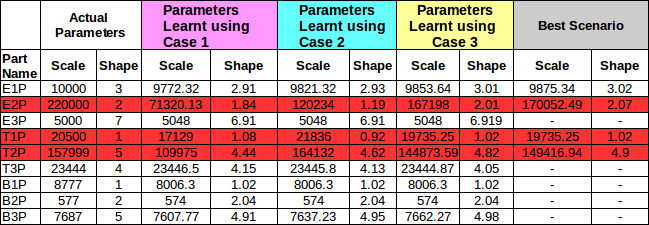}
\caption{Comparison of actual parameters with the parameters learnt using different approaches}
\label{Fig 1}
\end{figure}
\begin{figure}
\centering
\includegraphics[height=4.2cm]{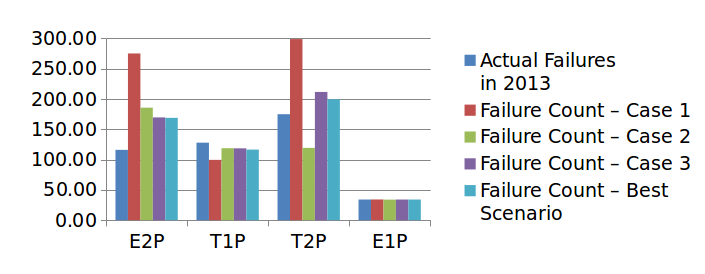}
\caption{Comparison of actual number of failures in year 2013 with the expected number  of failures calculated using different approaches}
\label{Fig 6}
\end{figure}
\begin{figure}
\centering
\includegraphics[height=4.2cm]{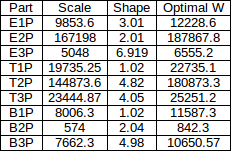}
\caption{Optimal warranty periods(in miles) for each part}
\label{Fig 1}
\end{figure}
\section{Related Work}
There are many studies which use statistical methods to integrate datasets from different sources, leveraging the interaction and correlation between them to obtain more refined information. In \cite{21}, joint likelihood model is used to combine gene expression and upstream sequence data for finding significant gene clusters. Similarly in \cite{22}, maximum likelihood method to predict protein-protein interactions and protein functions from three type of data. A kernel based data fusion is presented in \cite{20}. In this paper, we propose a model for fusion of data from multiple sources such as part failure data, tele-diagnostic data, service records using an approach based on Bayesian Network, which can learn the conditional dependency of symptom data on part-failure rate, resulting in improvement of warranty cost prediction accuracy. We also substantiate our claims through empirical results on simulated data, which has been inspired from real-life data. Bayesian Network has many real-world applications \cite{19}. In \cite{9}, Bayesian networks are used to analyze expression data. Similarly for cardiogenic heart failures, a continous time bayesian network model is presented in \cite{16}. Our approach based on Bayesian network has an application in warranty cost estimation of multi-component product family. We learn the dependence of symptom on part failure and predict number of future part failures with better accuracy.

Many studies use historical failure data to predict future failures in many studies \cite{2}, \cite{18}. In \cite{19} historical repair data is used to predict failure curves using Weibull distribution. In this paper, we propose an approach to use additional information like service records and tele-diagnostic data along with historical failure data to predict expected number of future failures with better accuracy.

\section{Conclusion and Future Work}
We have proposed an approach for fusion of information from diverse sources that can be used to enhance the accuracy of warranty cost estimates. We also described how our approach learns the dependencies of symptoms, gleaned from multiple sources of information, on actual failure rate of the parts. We have also shown that predicting future failures using these dependencies can help in better estimation of warranty costs. 
We have tested our approach on simulated data for the vehicles having past part-failure data and data from service stations and tele-diagnostics. We have shown that our approach can capture the information provided by the leading indicators and can predict future failures with better accuracy. Further, it was shown that our estimate of part-failure from symptom data is almost equal to actual part-failures, through comparison with the best case scenario. Hence, we conclude that failure rate analysis can be improved using service records and sensor data along with past part-failure data, using our approach.

In future, we plan to enhance the proposed model by incorporating more information such as geographical variations (hilly or plains areas) and seasonal variations. Currently, our model can predict expected number of failures with better accuracy as compared to traditional approach, which uses only part failure data. A possible next step our work is to extend our model to a framework that is capable of suggesting preventive maintenance in order to optimize the warranty cost. Also, the model needs to be tested on actual real-world datasets.

\end{document}